%% file: main.tex
\documentclass[runningheads]{llncs}

\usepackage{eccv}

\usepackage{eccvabbrv}

\usepackage{graphicx}
\usepackage{booktabs}

\usepackage[accsupp]{axessibility}  

\usepackage{hyperref}

\usepackage{orcidlink}

\begin{document}

\title{Reconstruction Matters: Learning Geometry-Aligned BEV Representation through 3D Gaussian Splatting} 

\titlerunning{Splat2BEV}

\author{Yiren Lu\inst{1,2} \and
Xin Ye\inst{1} \and
Burhaneddin Yaman \and
Jingru Luo\inst{1} \and
Zhexiao Xiong\inst{1,3} \and\\
Liu Ren\inst{1} \and
Yu Yin\inst{2}
}

\authorrunning{F.~Author et al.}

\institute{Bosch Research North America \&  Bosch Center for Artificial Intelligence (BCAI) \\
\and
Case Western Reserve University \\
\and
Washington University in St. Louis
}

\maketitle

\input{sec/0_abstract}
\input{sec/1_intro}

\input{sec/2_related_work}

\input{sec/3_method}
\input{sec/4_exp}

\input{sec/5_conclusion}


%
%
\bibliographystyle{splncs04}
\bibliography{main}
\end{document}

%% file: sec/0_abstract.tex
\begin{abstract}
Bird’s-Eye-View (BEV) perception serves as a cornerstone for autonomous driving, offering a unified spatial representation that fuses surrounding-view images to enable reasoning for various downstream tasks, such as semantic segmentation, 3D object detection, and motion prediction.
However, most existing BEV perception frameworks adopt an end-to-end training paradigm, where image features are directly transformed into the BEV space and optimized solely through downstream task supervision. 
This formulation treats the entire perception process as a black box, often lacking explicit 3D geometric understanding and interpretability, leading to suboptimal performance.
In this paper, we claim that an explicit 3D representation matters for accurate BEV perception, and we propose \textbf{Splat2BEV}, a Gaussian Splatting-assisted framework for BEV tasks. 
Splat2BEV aims to learn BEV feature representations that are both semantically rich and geometrically precise. 
We first pre-train a Gaussian generator that explicitly reconstructs 3D scenes from multi-view inputs, enabling the generation of geometry-aligned feature representations.
These representations are then projected into the BEV space to serve as inputs for downstream tasks.
Extensive experiments on nuScenes and argoverse dataset demonstrate that Splat2BEV achieves state-of-the-art performance and validate the effectiveness of incorporating explicit 3D reconstruction into BEV perception.
\end{abstract}

%% file: sec/1_intro.tex
\section{Introduction}
\label{sec:intro}

\begin{figure*}
    \centering
    \includegraphics[width=1\linewidth]{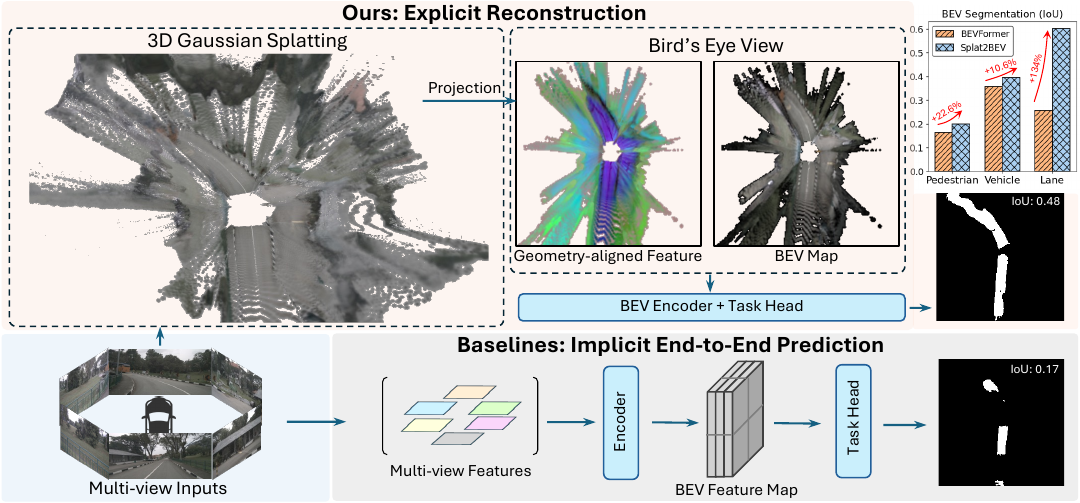}
    \caption{An overview of the proposed Splat2BEV framework.Compared to traditional implicit end-to-end approaches, Splat2BEV first reconstructs the scene using 3D Gaussian Splatting, and then projects the reconstructed scene into the Bird’s-Eye-View to obtain geometry-aligned representations for downstream tasks, leading to substantial performance improvements.}
    \label{fig:teaser}
\end{figure*}
Bird’s-Eye-View (BEV) perception has become a foundational basis for 3D scene understanding and autonomous driving.
By aggregating information from multiple surrounding views into a unified top-down representation, BEV perception allows consistent reasoning over object positions, motion trajectories, and environmental layouts.
Such a spatially structured representation serves as an effective bridge to downstream modules including segmentation \cite{philion2020lift, hu2021fiery, zhou2022cross}, detection \cite{li2024bevformer, liu2023bevfusion, xie2022m, huang2021bevdet} and planning \cite{dewangan2023uap, hu2023planning}.

Given that BEV serves as the backbone for autonomous driving tasks, numerous works have explored developing enhanced BEV representations. Existing BEV works can be broadly categorized into two main categories, namely 2D unprojection-based and 3D projection-based approaches.
2D unprojection-based methods \cite{liu2023bevfusion, philion2020lift} typically lift image features into the BEV plane through learned view transformations or depth-aware unprojection modules,
whereas 3D projection-based methods \cite{li2024bevformer, chambon2024pointbev} first construct feature representations within predefined 3D coordinate volumes and then project them into the BEV plane for downstream tasks. 

Despite the differences in methodological approaches to BEV generation, both lines of work follow an end-to-end training paradigm, in which BEV features are only learned through downstream task supervision. 
Such an implicit learning process for BEV features leads to the generation of imprecise and low-quality BEV feature maps that eventually results in suboptimal performance for critical downstream autonomous driving tasks~\cite{le2024diffuser,ye2025bevdiffuser}. While more recent works have tried to introduce additional supervision~\cite{ye2025bevdiffuser}, they still fail to address the core issue, the lack of explicit representation, as they do not enforce a geometry-aligned BEV reconstruction capable of producing highly precise and accurate BEV feature maps.


In this paper, we propose \textit{Splat2BEV}, a Gaussian Splatting-assisted framework for BEV perception. 
\textit{Splat2BEV} incorporates 3D Gaussian Splatting as an explicit intermediate representation within the BEV perception pipeline.
Leveraging 3D reconstruction, our framework generates geometry-aligned feature representations that enhance the effectiveness of downstream BEV tasks.
An overview of the proposed pipeline and its comparison with the conventional end-to-end paradigm are illustrated in \cref{fig:teaser}.
Given perspective-view images captured by vehicle-mounted cameras, 
we first pre-train a Gaussian generator to \textit{explicitly} reconstruct the 3D scene from multi-view inputs, during which semantic knowledge from vision foundation models is distilled into the reconstructed representation.
This process enables the learning of a geometry- and semantics-aligned feature field that captures both structural and contextual information.
Leveraging the differentiable rendering capability of 3D Gaussian Splatting, the reconstructed Gaussians and their associated features can be rasterized from a top-down viewpoint to form a BEV feature map, serving as input for downstream tasks.
Such explicit reconstruction and projection promote the learning of geometrically consistent and semantically meaningful BEV features, enhancing both interpretability and representation quality.
Finally, a lightweight segmentation head operates on the BEV feature map to produce pixel-wise predictions for applications such as vehicle, pedestrian, and lane segmentation. 
Extensive experiments on the nuScenes dataset show that \textit{Splat2BEV} significantly outperforms existing baselines, achieving a 21.4\% improvement in lane segmentation and an average gain of 11.0\% across vehicle, pedestrian, and lane classes, highlighting the advantage of explicit reconstruction over implicit representations.
Additional multi-class segmentation results on both nuScenes and Argoverse1 further demonstrate the robustness of our method.

In summary, the main contribution of our paper includes:
\begin{itemize}
    \item We propose \textit{Splat2BEV}, a novel Gaussian Splatting-assisted framework for BEV perception, showcasing the importance of explicit reconstruction in enhancing BEV perception for the first time.
    \item We incorporate vision foundation models to inject fine-grained visual cues into the reconstruction, improving the quality and expressiveness of the BEV feature representation.
    \item Extensive quantitative and qualitative experiments on the nuScenes and argoverse1 dataset  show that \textit{Splat2BEV} achieves the state-of-the-art performance and consistently outperforms the baseline methods.
\end{itemize}

%% file: sec/2_related_work.tex
\section{Related Work}
\label{sec:related_work}
\subsection{BEV Perception}
BEV provides a unified space for integrating information from multiple cameras, making it one of the primary representations used in autonomous driving.
A crucial step in BEV perception lies in transforming features from the perspective camera space into the BEV space.
In this section, we categorize existing transformation approaches into three main groups: \textit{learning-based unprojection}, \textit{geometry-based unprojection} and \textit{projection from learned 3D volumes}.

\noindent\textbf{Geometry-based unprojection.}
Geometry-based approaches explicitly lift 2D image coordinates into 3D space.
Lift-Splat-Shoot (LSS) \cite{philion2020lift} predicts per-pixel depth and uses camera intrinsics and extrinsics to map image features to 3D points through a geometric unprojection process.
However, its performance heavily depends on the quality of the predicted depth.
Since the training process relies solely on downstream task supervision, the depth estimation can be unstable or inaccurate.
Subsequent works \cite{li2023bevstereo, li2023bevdepth, xie2022m, yang2023parametric} extend the LSS paradigm by introducing additional depth supervision to improve the accuracy and robustness of depth prediction.
BEVDepth \cite{li2023bevdepth} leverages LiDAR-derived depth maps for supervision, while BEVStereo \cite{li2023bevstereo} employs stereo matching techniques to enhance depth estimation quality.

\noindent\textbf{Learning-based unprojection.}
Learning-based methods \cite{liu2022petr, liu2023petrv2, pan2023baeformer, zhou2022cross, bartoccioni2023lara} implicitly learn the mapping from 2D to 3D coordinates in a data-driven manner. 
These approaches employ attention-based architectures or multi-layer perceptrons (MLPs) to directly associate image features with 3D spatial locations.
Instead of explicitly predicting depth, they encode 3D positional information into learnable queries and use cross-attention to aggregate multi-view image features into BEV space.
However, these methods remain fully implicit, lacking geometric reasoning and interpretability.

\noindent\textbf{Projection from learned 3D volumes.}
Instead of directly lifting image features or learning implicit mappings, another category of methods constructs a predefined 3D volume and learns dense feature representations within it.
BEVFormer \cite{li2024bevformer} and \cite{harley2023simple} learn a 3D feature grid that captures spatial structure and semantics across the scene.
To obtain BEV features, these methods project or collapse the learned 3D feature volume onto the BEV plane, generating top-down representations.
PointBEV \cite{chambon2024pointbev} further introduce a coarse-to-fine training strategy that focuses on regions of interest to reduce memory consumption

However, these methods are trained solely with downstream task losses without explicit 3D reconstruction, leading to a lack of interpretability.
In contrast, we propose Splat2BEV, which introduces explicit 3D reconstruction during both training and inference, and projects the learned representation into BEV space for downstream tasks.

\subsection{Gaussian Splatting in Autonomous Driving}
Recently, a growing number of studies have incorporated Gaussian Splatting into the autonomous driving tasks.

\noindent\textbf{Navigation.}
Several recent works \cite{keetha2024splatam, matsuki2024gaussian, yan2024gs, huang2024photo, peng2024rtg} integrate 3D Gaussian Splatting into robot navigation and SLAM systems as the mapping component.
They extend traditional 3D Gaussian Splatting to support online reconstruction, enabling continuous scene updates during exploration.
These approaches demonstrate that Gaussian primitives can provide a compact and differentiable representation for mapping in autonomous driving scenarios.

\noindent\textbf{Simulation.}
Another line of research \cite{yan2024street, zhou2024drivinggaussian, zhou2024hugs, hess2025splatad} leverages 3D Gaussian Splatting to reconstruct realistic environments for autonomous driving simulation.
They employ Gaussian Splatting to build photorealistic and geometrically consistent driving scenes from real-world data.
By transforming large-scale outdoor environments into editable Gaussian representations, these approaches enable efficient rendering, controllable scene manipulation, and sensor simulation for training and evaluating perception and planning models in autonomous driving.

\noindent\textbf{BEV Perception.}
GaussianLSS \cite{lu2025toward} and GaussianBEV \cite{chabot2025gaussianbev} introduces 3D Gaussian Splatting as an intermediate representation, enabling a more efficient aggregation of BEV feature.
However, these two Gaussian Splatting–based methods treat the Gaussians merely as feature carriers rather than performing explicit 3D scene reconstruction.
Their learning process still relies solely on downstream task supervision, limiting geometric interpretability.

%% file: sec/3_method.tex
\section{Methodology}
\label{sec:method}
\begin{figure*}[htbp]
    \centering
    \includegraphics[width=1\linewidth]{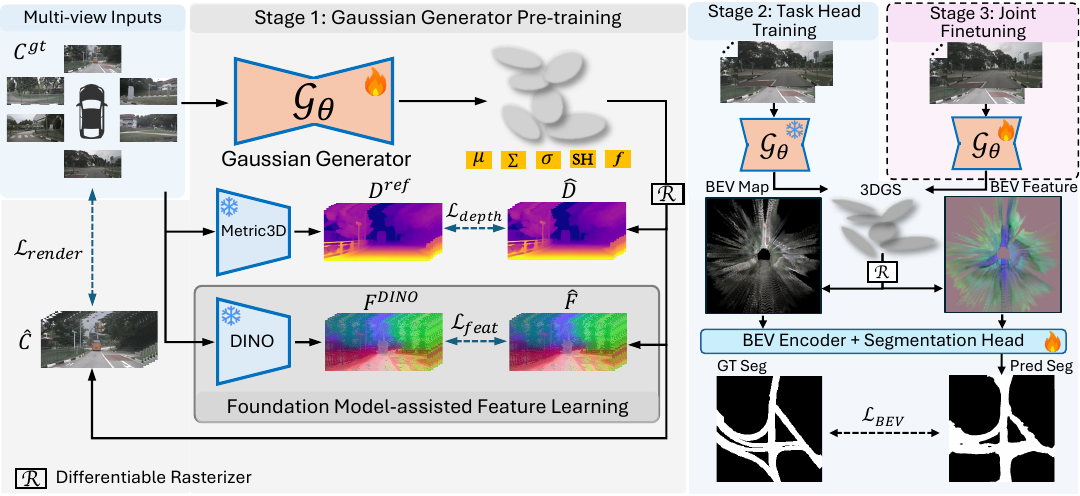}
    \caption{
    \textbf{An overview of our training process.}
    Given multi-view perspective images as input, Splat2BEV first trains a feed-forward Gaussian generator to reconstruct 3D scene using 3D Gaussian Splatting.
    In stage 2, the Gaussian generator is frozen, and the reconstructed geometry along with its associated features are projected onto the BEV plane.
    A BEV encoder and segmentation head are then trained on top of this BEV representation to perform downstream tasks.
    Finally, in the third stage, the Gaussian generator, BEV encoder, and segmentation head are jointly fine-tuned, allowing geometry, semantics, and task-specific cues to be harmonized for optimal BEV perception.
    }
    \label{fig:training_pipeline}
\end{figure*}

Our proposed Splat2BEV is a Gaussian Splatting–assisted framework that explicitly reconstructs the scene for accurate Bird’s-Eye-View (BEV) perception.
An overview of our framework and training process is illustrated in \cref{fig:training_pipeline}.
Given multi-view perspective images as input,
Splat2BEV first pre-trains a feed-forward Gaussian generator capable of reconstructing 3D scenes (\cref{subsec:GS_generator} \& \cref{subsec:feature_learning}).
In the second stage, the Gaussian generator is frozen, and the reconstructed scenes are projected into the Bird’s-Eye-View (BEV) for downstream task head training (\cref{subsec:bev_projection}, \cref{subsec:task_head}).
In the final stage, all components are jointly fine-tuned to harmonize geometric, semantic, and task-specific cues for optimal BEV perception (\cref{subsec:joint_finetuning}).
We detail each stage in the following sections.

\subsection{Preliminary: 3D Gaussian Splatting}
In Splat2BEV, we adopt 3D Gaussian Splatting (3DGS) as the scene representation. 
3DGS models a scene using a set of anisotropic Gaussians $\{G_i\}_{i=1}^n$, where each Gaussian $G_i = (\boldsymbol{\mu}_i, \boldsymbol{\Sigma}_i, \boldsymbol{\sigma}_i, \mathbf{SH}_i)$ is characterized by its mean  $\boldsymbol{\mu}_i$, covariance $\boldsymbol{\Sigma}_i$, 
opacity $\boldsymbol{\sigma}_i$, and a set of spherical harmonics coefficients $\mathbf{SH}_i$ that encode view-dependent color. 

A 3DGS-represented scene can be rendered from arbitrary viewpoints through a differentiable splatting process, 
in which 3D Gaussians are projected onto the image plane as 2D elliptical footprints, and their colors are accumulated to form pixel intensities. 
Specifically, for each pixel $p$, the final rendered color $\mathbf{C}(p)$ is computed through $\alpha$-blending:
\begin{equation}
    \mathbf{C}(p) = \sum_{i=1}^{N} T_i \, \sigma_i \, \mathbf{c}_i, 
    \  \text{where} \ 
    T_i = \prod_{j < i} (1 - \sigma_j).
    \label{eq:alpha_blending}
\end{equation}
Here, $\mathbf{c}_i$ denotes the view-dependent color decoded from $\mathbf{SH}_i$, 
and $\sigma_i$ represents the opacity of the Gaussians after depth sorting. 
Leveraging this ability, we can utilize 3DGS to directly synthesize the reconstructed scene from a top-down perspective to obtain the BEV representation.

\subsection{Gaussian Generator Pre-training}
\label{subsec:GS_generator}
\begin{figure}
    \centering
    \includegraphics[width=0.65\linewidth]{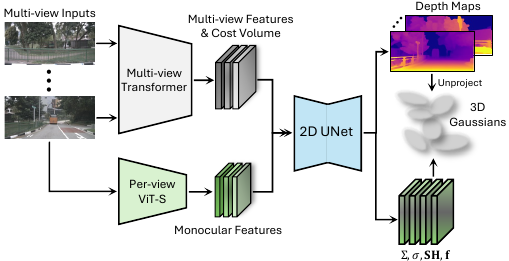}
    \caption{Our Gaussian generator consists of two branches: a multi-view branch based on UniMatch \cite{xu2023unifying}, and a per-view branch built upon ViT-S \cite{dosovitskiy2020vit} to augment the multi-view representation.
    The multi-view branch outputs multi-view features and cost volumes, which are concatenated with monocular features from the per-view branch to serve as input for a U-Net \cite{ronneberger2015u} to predict depth maps and per-Gaussian parameters.
    The predicted depths are then unprojected and combined with the Gaussian parameters to form 3D Gaussians.}
    \label{fig:GS_generator}
    \vspace{-1em}
\end{figure}

Given a set of $k$ perspective images $\mathbf{I} = \{I_i\}_{i=1}^k$ as input, we train a Gaussian generator $\mathcal{G}_\theta$ to output a set of Gaussians $\mathbf{G} = \{G_j\}_{j=1}^{n}$ to represent the scene. 
Following previous feed-forward Gaussian Splatting methods~\cite{charatan2024pixelsplat,chen2024mvsplat,xu2025depthsplat}, 
$\mathcal{G}_\theta$ first predicts a depth map for each view, which is then unprojected into 3D space using the corresponding camera parameters to obtain the mean positions of the Gaussians. 
Subsequently, separate prediction heads are employed to estimate additional Gaussian attributes (\ie, opacity, covariance and color).
Detailed structure of the Gaussian generator is shown in \cref{fig:GS_generator}.
These outputs are organized into 3D Gaussians and rendered back into the input views for supervision using MSE loss:
\begin{equation}
    \mathcal{L}_{\text{render}} =
    \frac{1}{k} \sum_{i=1}^{k} 
    \| \widehat{\mathbf{C}}_i - \mathbf{C}_i^{\text{gt}} \|_2^2,
\end{equation}
\begin{equation}
    \widehat{\mathbf{C}}_i = 
    \mathcal{R}\big(\mathbf{G}; \mathbf{P}_i\big).
\end{equation}
Here, $\mathcal{R}(\cdot)$ denotes the differentiable rendering function that synthesizes the image 
from the reconstructed Gaussians $\mathbf{G}$ given the camera pose $\mathbf{P}_i$.

Given the surrounding view cameras on autonomous vehicles are designed to maximize coverage of the environment and exhibits limited overlap region between adjacent frames,  reconstruction based solely on the rendering loss becomes highly challenging. 
To further assist the reconstruction process, we employ the depth foundation model Metric3Dv2 \cite{hu2024metric3d} to generate reference metric depth for additional supervision. 
The predicted depth $\widehat{\mathbf{D}}_i$ is computed by accumulating the camera-space depth ($z$) values of all projected Gaussians along the viewing ray.
For depth supervision,
we first apply an L1 loss to enforce pixel-wise metric depth accuracy:
\begin{equation}
    \mathcal{L}_{\text{depth}}^{L1} = 
    \frac{1}{k} \sum_{i=1}^{k} 
    \| \widehat{\mathbf{D}}_i - \mathbf{D}_i^{\text{ref}} \|_1,
\end{equation}
where $\widehat{\mathbf{D}}_i$ and $\mathbf{D}_i^{\text{ref}}$ denote the predicted and reference depth maps for the $i$-th view, respectively. 
Besides, to ensure scale-invariant geometric consistency, we further employ the SILog loss \cite{eigen2014depth}, defined as:
\begin{equation}
    \mathcal{L}_{\text{depth}}^{\text{SILog}} = 
    \frac{1}{k} \sum_{i=1}^{k} 
    \Bigg[
        \frac{1}{n} \sum_{p} \delta_i(p)^2
        - \frac{1}{n^2} 
        \Big( \sum_{p} \delta_i(p) \Big)^2
    \Bigg],
\end{equation}
\begin{equation}
    \delta_i(p) = 
    \log \widehat{\mathbf{D}}_i(p) - \log \mathbf{D}_i^{\text{ref}}(p),
\end{equation}
where $\delta_i(p)$ denotes the per-pixel log-depth difference and $n$ is the number of valid pixels. 
\subsection{Foundation Model-assisted Feature Learning}
\label{subsec:feature_learning}
Most of the existing BEV perception models solely rely on downstream task loss to provide supervision signal, leaving intermediate BEV features weakly constrained and not guaranteed to be geometrically or semantically meaningful.
To address this issue and inject rich semantic priors, we attach a learnable feature vector $\mathbf{f}_j \in \mathbb{R}^{C}$ to each Gaussian $G_j$ and distill dense visual cues from a frozen DINO~\cite{oquab2024dinov2} encoder.
Formally, the Gaussian generator is defined as:
\begin{equation}
\begin{gathered}
    \mathcal{G}_\theta(\mathbf{I}) = \mathbf{G},\\[2pt]
    \text{where }\mathbf{G} = \{\, G_j \mid G_j = (\boldsymbol{\mu}_j, \boldsymbol{\Sigma}_j, \boldsymbol{\sigma}_j, \mathbf{SH}_j, \mathbf{f}_j) \,\}.
\end{gathered}
\end{equation}
The rendering of feature also follows \cref{eq:alpha_blending} by substituting color $\mathbf{c}$ with feature vector $\mathbf{f}$.
For feature supervision, we employ a cosine similarity loss between the rendered feature map $\widehat{\mathbf{F}}$ 
and the dense feature map extracted from the frozen DINO encoder $\mathbf{F}^{\text{DINO}}$:
\begin{equation}
    \mathcal{L}_{\text{feat}} = 
    \frac{1}{k} \sum_{i=1}^{k} 
    \Big( 1 - 
    \cos\big( 
    \widehat{\mathbf{F}}_i, \,
    \mathbf{F}^{\text{DINO}}_i
    \big) \Big).
\end{equation} 
This distillation aligns each Gaussian feature vector $\mathbf{f}_j$ with the high-level semantics extracted by DINO, yielding geometrically grounded and semantically meaningful features.

The overall loss function used to train the Gaussian generator $\mathcal{G}_\theta$ is defined as:
\begin{equation}
    \mathcal{L}_{\text{total}} = 
    \mathcal{L}_{\text{render}} 
    + \lambda_1 \, \mathcal{L}_{\text{depth}}^{L1} 
    + \lambda_2 \, \mathcal{L}_{\text{depth}}^{\text{SILog}} + \lambda_3 \, \mathcal{L}_{\text{feat}},
    \label{eq:loss_total}
\end{equation}
where $\lambda_1$, $\lambda_2$ and $\lambda_3$ are weighting coefficients that balance the photometric, depth and feature supervision terms. 

\subsection{Projection to BEV Space}
\label{subsec:bev_projection}
After the Gaussians are generated, we project them onto the Bird’s-Eye-View (BEV) plane to serve as inputs for subsequent BEV-based tasks. 
Unlike the perspective projection commonly used in Gaussian Splatting for novel-view rendering, 
the BEV projection adopts an \textit{orthogonal projection} to preserve geometric consistency in the top-down view. 
The orthogonal projection matrix $\mathbf{M}_{\text{ortho}}$ in camera space and the projection of Gaussian means are defined as:
\begin{align}
\mathbf{M}_{\text{ortho}} &=
\begin{bmatrix}
\label{eq:ortho_proj}
f_x & 0 & 0 & c_x \\
0 & f_y & 0 & c_y \\
0 & 0 & 0 & 1
\end{bmatrix}, \\
\begin{bmatrix}
u \\ v \\ 1
\end{bmatrix}
&=
\mathbf{M}_{\text{ortho}}
\begin{bmatrix}
G_X \\ G_Y \\ G_Z \\ 1
\end{bmatrix}.
\end{align}
Here, $f_x$, $f_y$, $c_x$, $c_y$ denote the BEV camera parameters; $G_X$, $G_Y$, $G_Z$ represent the 3D Gaussian mean in camera-space coordinates; and $u$, $v$ represent the 2D Gaussian mean on the image plane.
Since the orthogonal projection is a linear transformation, 
the projection of the 3D covariance matrix $\boldsymbol{\Sigma}_{3D}$ is straightforward. 
From \cref{eq:ortho_proj}, we obtain the Jacobian $\mathbf{J}_{\text{ortho}}$ of $\mathbf{M}_{\text{ortho}}$ as follows:
\begin{equation}
\mathbf{J}_{\text{ortho}}
=
\begin{bmatrix}
f_x & 0 & 0 \\
0 & f_y & 0
\end{bmatrix},
\end{equation}
and the projection of the 3D Gaussian covariance into 2D is defined as:
\begin{equation}
\boldsymbol{\Sigma}_{2D}
= 
\mathbf{J}_{\text{ortho}}
\, \boldsymbol{\Sigma}_{3D} \,
\mathbf{J}_{\text{ortho}}^{\!\top},
\end{equation}
where $\boldsymbol{\Sigma}_{2D}$ is the projected 2D covariance matrix on the image plane. 
The orthogonal projection of 3D Gaussians is implemented using \texttt{gsplat}~\cite{ye2025gsplat}.

\subsection{Task Head Training}
\label{subsec:task_head}
After projecting the reconstructed geometry and associated features into BEV space, we obtain a geometry-aligned BEV representation.
To produce task-specific BEV features, this representation is fed into a BEV encoder.
A downstream segmentation head is then attached, taking the output of the BEV encoder to perform BEV segmentation.
During this stage, the Gaussian generator is kept frozen. Only the parameters of the BEV encoder and the task head are optimized, enabling the network to adapt the learned respresentation for downstream tasks without altering the underlying 3D reconstruction.

Following previous works in BEV segmentation, we train the BEV encoder and segmentation head using a combination of sigmoid focal loss~\cite{lin2017focal}, centerness L2 loss, and offset L1 loss:
\begin{equation}
    \mathcal{L}_{\text{BEV}} = \mathcal{L}_{\text{focal}} + \lambda_4 \, \mathcal{L}_{\text{center}} + \lambda_5 \, \mathcal{L}_{\text{offset}}.
    \label{eq:bev_loss}
\end{equation}
This multi-term loss encourages accurate segmentation masks while maintaining spatial alignment between predicted centers and offsets in the BEV domain.

\subsection{Joint Fine-tuning}
\label{subsec:joint_finetuning}
After task head–only training, Splat2BEV already possesses the ability to perform BEV segmentation and achieves considerable performance, refer to \cref{tab:frozen_backbone_finetuning} for detailed results.
To further harmonize geometry, semantics, and task-specific representations, we jointly fine-tune the Gaussian generator, BEV encoder, and segmentation head in an end-to-end manner.
This stage enables gradients from the BEV segmentation loss to flow back into the Gaussian generator, allowing the reconstructed 3D scene and feature field to adapt to task-level objectives. 
The overall loss function remains identical to \cref{eq:bev_loss}.

%% file: sec/4_exp.tex
\section{Experiments}
\label{sec:exp}
\subsection{Dataset and Evaluation Metric}
\noindent\textbf{Dataset.}
We evaluate Splat2BEV on the nuScenes dataset \cite{caesar2020nuscenes} and the argoverse1 dataset \cite{chang2019argoverse}.
NuScenes is a comprehensive benchmark for autonomous driving, including 1000 driving scenes collected in Boston and Singapore under diverse weather and lighting conditions, with 700 scenes used for training, 150 for validation, and 150 for testing. 
Argoverse 1 is another large-scale autonomous driving dataset collected in Pittsburgh and Miami. 
It contains 113 driving logs with synchronized multi-view camera images and LiDAR point clouds.
All training is conducted on the training split, and all quantitative results are reported on the validation set.
We first conduct experiments on three categories separately on nuScenes, \ie, \textit{vehicle}, \textit{pedestrian}, and \textit{lane} segmentation, following the settings of \cite{chambon2024pointbev, lu2025toward}.
We then conduct multi-class BEV segmentation experiments on both nuScenes and Argoverse 1, following the settings of \cite{zou2024diffbev, zhao2024improving}.

\noindent\textbf{Metric.}
We report the Intersection-over-Union (IoU) as the primary evaluation metric across all tasks, following the well-established practice of prior works. 

\subsection{Implementation Details}
\noindent\textbf{Input Image.}
In our experiments, the multi-view input images are resized to either 224 $\times$ 480 or 448 $\times$ 800 resolution, depending on the experiment setting.
Unless otherwise specified, the default input resolution is 224 $\times$ 480.

\noindent\textbf{Loss Coefficient.}
For the loss term coefficients, in $\mathcal{L}_{\text{total}}$ (\cref{eq:loss_total}), we empirically set $\lambda_1 = 0.2$, $\lambda_2 = 0.8$, and $\lambda_3 = 0.1$. 
In $\mathcal{L}_{\text{BEV}}$ ( \cref{eq:bev_loss}), we set $\lambda_4 = 2.0$ and $\lambda_5 = 0.1$.

\noindent\textbf{BEV Projection.}
In the BEV projection stage, 
the BEV representation has a spatial resolution of $200 \times 200$, and the BEV feature map output by the BEV encoder maintains the same resolution. 
The BEV focal lengths are set to $f_x = f_y = 2$, computed as the ratio between the BEV pixel resolution (200) and the BEV spatial range (100\,m). 
The pixel offsets are set to $c_x = c_y = 100$, corresponding to half of the BEV pixel resolution.
For the extrinsics, the BEV camera is positioned at $+3$\,m along the $z$-axis in the vehicle coordinate frame.

\noindent\textbf{Training Strategy.}
During Gaussian generator pre-training, learning rate of the multi-view branch is set to $2\times10^{-4}$.
For the monocular branch, we initialize from the pretrained Depth Anything V2 \cite{depth_anything_v2} weights and adopt a smaller learning rate of $2\times10^{-6}$ to preserve pretrained knowledge.
In stage 2, we freeze the Gaussian generator and only optimize BEV encoder and segmentation head with a learning rate of $2\times10^{-4}$.
During joint finetuning, the learning rate of the monocular branch remains $2\times10^{-6}$, while all other components are trained with a learning rate of $2\times10^{-6}$. 
We pre-train the Gaussian generator for 10 epochs, perform task head-only finetuning for 10 epochs, and finally joint finetuning for an additional 20 epochs.

\subsection{Quantitative Comparison}
We mainly conduct experiment on three key tasks: vehicle segmentation, pedestrian and lane segmentation. We compare the performance of Splat2BEV with a collection of BEV segmentation methods.

\noindent\textbf{Vehicle Segmentation.}
Following the evaluation setting of previous works \cite{chambon2024pointbev, kerbl20233d}, we utilize two different input resolutions: 224x480 and 448x800. 
And we also include experiments with visibility filtering (\ie only consider objects with $> 40\%$ visibility).
The quantitative results for vehicle segmentation is shown in \cref{tab:vehicle_segmentation}. 
Splat2BEV outperforms all the baseline methods under various experiment settings.
\input{tab/vehicle_seg}

\noindent\textbf{Pedestrian Segmentation.}
For pedestrian segmentation, following the evaluation protocol of previous works, we consider only pedestrians with visibility $>40\%$.
The comparison results are shown in \cref{tab:pedestrian_segmentation}.
Splat2BEV achieves an 8\% improvement compared to baseline methods.
Since pedestrians occupy smaller regions compared to vehicles, accurate 2D-3D correspondence is critical, even minor projection errors can lead to completely non-overlapping segmentation results.
The superior performance of our method demonstrates its ability to achieve more accurate geometric mapping through explicit 3D reconstruction.

\noindent\textbf{Lane Segmentation.}
The quantitative results for lane segmentation are presented in \cref{tab:lane_segmentation}.
Splat2BEV achieves a 21.4\% improvement over previous methods.
Lanes are one of the largest and most structured components in autonomous driving scenes, making lane segmentation a task that particularly benefits from explicit 3D reconstruction.
\input{tab/combined_ped_lane}

\noindent\textbf{Multi-class Segmentation}
We then evaluate Splat2BEV on multi-class segmentation on both nuScenes and argoverse1 to showcase the generalization ability and robustness of our method, the results are show in \cref{tab:multiclass_nuscenes} and \cref{tab:multiclass_argoverse}.
\input{tab/multiclass_nuscenes}
\input{tab/multiclass_argoverse}

Overall, Splat2BEV consistently outperforms existing BEV segmentation frameworks across all evaluated tasks. 
These improvements highlight the effectiveness of incorporating explicit 3D reconstruction to learn geometry-aligned representations. 
Such representations enable Splat2BEV to achieve more accurate and interpretable feature alignment, providing benefits for both small-scale objects (\ie, pedestrians) and large-scale structural elements (\ie, lanes).
\begin{figure*}
    \centering
    \includegraphics[width=1\linewidth]{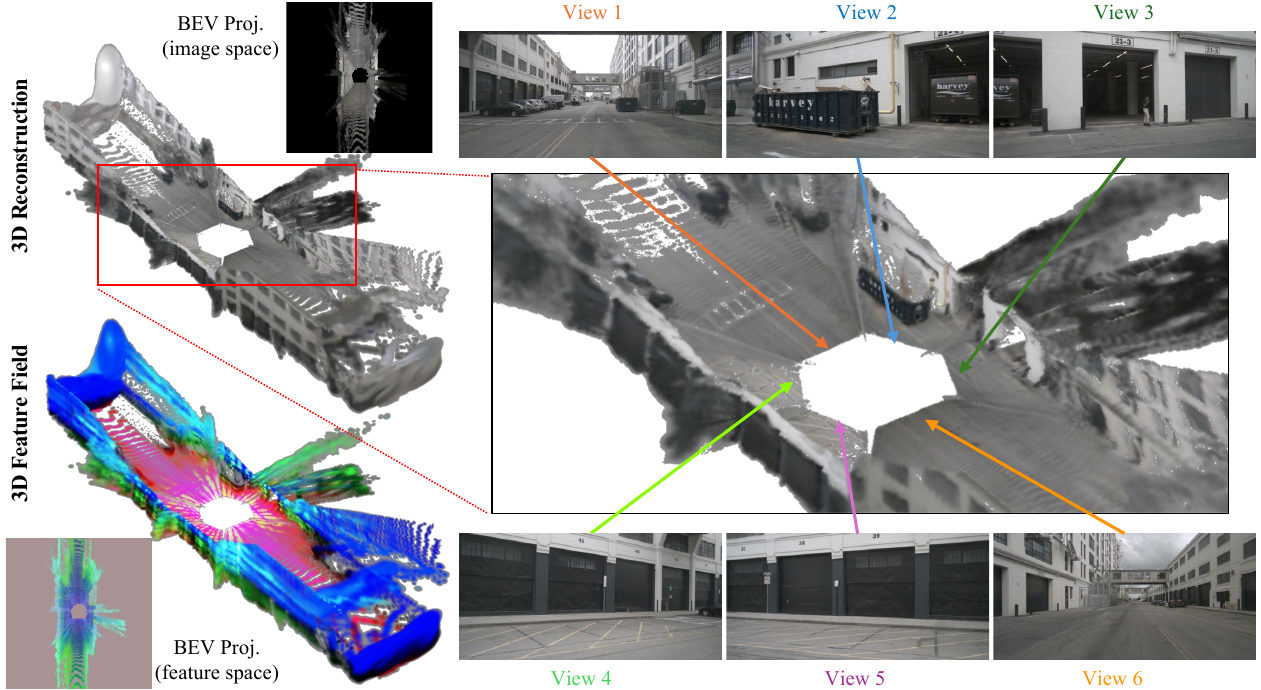}
    \vspace{-1em}
    \caption{
    \textbf{Visualization of reconstruction quality.}
    The left side shows the 3D reconstruction, its feature field, and the corresponding BEV map and projected BEV feature.
    The right side provides zoomed-in views that highlight fine-grained details, such as the zebra crossing in View~1, the dustbin in View~2, the person in View~3, and the no-parking line in View~4.
}
    \label{fig:reconstruction}
    \vspace{-1em}
\end{figure*}
\begin{figure*}
    \centering
    \includegraphics[width=1\linewidth]{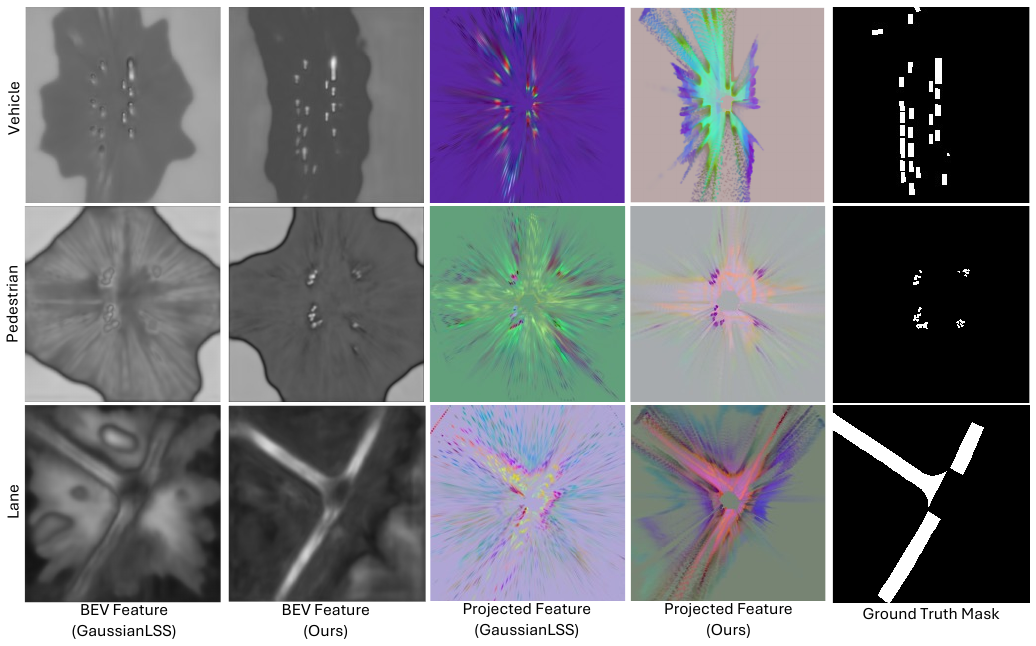}
    \vspace{-1.5em}
    \caption{\textbf{Visual comparison of features learned with and without explicit reconstruction.} 
    The \textit{BEV feature} refers to the feature map produced by the BEV encoder, while the \textit{projected feature} denotes the feature directly projected from the 3D representation.}
    \label{fig:feature_vis}
\end{figure*}
\subsection{Qualitative Results on Reconstruction}
To qualitatively evaluate the reconstruction quality of our Gaussian generator, we visualize a reconstructed scene in \cref{fig:reconstruction}. Additional visualizations are provided in the supplementary materials. 
On the left side, we show the reconstructed 3D scene together with its corresponding feature field. 
On the right side, we highlight several fine-grained details in the reconstruction to further demonstrate the fidelity and quality of the recovered scene.

\subsection{Analysis of Learned BEV Representation}
\input{tab/frozen_backbone_finetune}
To examine the quality of the features learned during Gaussian generator pre-training (\ie, \cref{subsec:GS_generator}), we directly evaluate the performance of the Splat2BEV without joint fine-tuning, where the Gaussian generator is kept frozen.
The quantitative results are presented in \cref{tab:frozen_backbone_finetuning}.
Surprisingly, even without joint optimization, our method already attains performance comparable to several baselines, indicating that the geometry-aligned features produced by Gaussian generator pre-training and feature distillation are both meaningful and of high quality. 
After enabling joint fine-tuning, the performance further improves, validating the benefits of integrating the learned explicit 3D representation into downstream BEV segmentation.
Since our method incorporates an explicit reconstruction process, the learned features are naturally aligned with the underlying scene geometry.
We further visualize the BEV feature maps in \cref{fig:feature_vis}.
For comparison, we select GaussianLSS \cite{lu2025toward}, another 3DGS-based framework, to highlight how explicit reconstruction leads to more structured and geometry-aware BEV features.
We visualize both the features obtained by directly projecting the 3D representation and the BEV feature maps produced by the BEV encoder.

\subsection{Ablation Study}

We conduct ablation studies on two design strategies: the proposed 3-stage training scheme (compared with a 2-stage variant without \cref{subsec:task_head}), and DINO feature distillation. 
The results are shown in \cref{tab:ablation_module}.

Since our method explicitly reconstructs and projects the scene into a virtual BEV space, the choice of BEV camera height is crucial. 
Placing the BEV camera too low may lead to missing geometry or objects in the scene, whereas placing it too high may introduce extraneous distractors (\eg, trees and poles) that harm downstream task performance.
Therefore, we conduct an ablation study in \cref{tab:height_ablation} to investigate the impact of different BEV camera heights on overall perception accuracy.
Our results show that setting the virtual BEV camera height to 3 meters achieves the best trade-off, effectively capturing essential ground-level geometry while suppressing irrelevant high-altitude structures.
\input{tab/combined_ablation}

%% file: tab/vehicle_seg.tex
\begin{table}[t]
    \centering
    \caption{\textbf{Comparison of BEV segmentation results for Vehicle on the nuScenes dataset.}}
    {\setlength{\tabcolsep}{5pt}
    \resizebox{0.7\linewidth}!{%
    \begin{tabular}{lcccccc}
        \toprule
        {\textbf{Vehicle IoU} ($\uparrow$)} & \multicolumn{2}{c}{\textbf{No visibility filtering}} & \multicolumn{2}{c}{\textbf{Visibility filtering}} \\
        \textbf{Method} & \textbf{224 $\times$ 480} & \textbf{448 $\times$ 800} & \textbf{224 $\times$ 480} & \textbf{448 $\times$ 800} \\
        \hline
        BEVFormer \cite{li2024bevformer} & 35.8 & 39.0 & 42.0 & 45.5 \\
        Simple-BEV \cite{harley2023simple} & 36.9 & 40.9 & 43.0 & 44.9 \\
        PointBeV \cite{chambon2024pointbev} & 38.7 & 42.1 & 44.0 & 47.6 \\
        CVT \cite{zhou2022cross} & 31.4 & 32.5 & 36.0 & 37.7 \\
        BAEFormer \cite{pan2023baeformer} & 36.0 & 37.8 & 38.9 & 41.0 \\
        GaussianLSS \cite{lu2025toward} & 38.3 & 40.6 & 42.8 & 46.1 \\
        Ours & \textbf{39.6} & \textbf{42.7} & \textbf{44.6} & \textbf{48.2} \\
        \bottomrule
    \end{tabular}}
    }
\label{tab:vehicle_segmentation}
\end{table}

%% file: tab/combined_ped_lane.tex
\begin{table}[t]
\centering
\small

\begin{minipage}[htbp]{0.45\linewidth}
\centering
\caption{\textbf{BEV pedestrian segmentation on nuScenes.}}
\vspace{-0.5em}
{\setlength{\tabcolsep}{6pt}
\begin{tabular}{lc}
\toprule
\textbf{Pedestrian segmentation} & \textbf{IoU (\(\uparrow\))} \\
\midrule
BEVFormer \cite{li2024bevformer} & 16.4 \\
SimpleBEV \cite{harley2023simple} & 17.1 \\
PointBeV \cite{chambon2024pointbev} & 18.5 \\
LSS \cite{philion2020lift} & 15.0 \\
CVT \cite{zhou2022cross} & 14.2 \\
GaussianLSS \cite{lu2025toward} & 18.0 \\
Ours & \textbf{20.1} \\
\bottomrule
\end{tabular}}
\label{tab:pedestrian_segmentation}
\end{minipage}
\hfill
\begin{minipage}[htbp]{0.48\linewidth}
\centering
\caption{\textbf{BEV lane segmentation on nuScenes.}}
{\setlength{\tabcolsep}{6pt}
\begin{tabular}{lc}
\toprule
\textbf{Lane segmentation} & \textbf{IoU (\(\uparrow\))} \\
\midrule
BEVFormer \cite{li2024bevformer} & 25.7 \\
PETRv2 \cite{liu2023petrv2} & 44.8 \\
M2BEV \cite{philion2020lift} & 38.0 \\
MatrixVT \cite{zhou2022cross} & 44.8 \\
PointBeV \cite{chambon2024pointbev} & 49.6 \\
Ours & \textbf{60.2} \\
\bottomrule
\end{tabular}}
\label{tab:lane_segmentation}
\end{minipage}
\vspace{-1em}
\end{table}

%% file: tab/multiclass_nuscenes.tex
\begin{table}[htbp] 
 \centering
 \scriptsize
 \setlength{\tabcolsep}{2pt} 
 \caption{\textbf{Multi-class segmentation results on nuScenes.}}
 \scalebox{0.93}{
 \begin{tabular}{ccccccccccccc}
    \toprule
    \textbf{Method} & \textbf{Driv.} & \textbf{Walk.} & \textbf{Cross.} & \textbf{Carpark} & \textbf{Car} & \textbf{Truck} & \textbf{Bus} & \textbf{Trailer} & \textbf{C.V.} & \textbf{Ped.} & \textbf{Motor.} & \textbf{Mean}\\
    \midrule
    VPN~\cite{pan2020cross}    & 58.0 & 29.4 & 27.3 & 12.9 & 25.5 & 17.3 & 20.0 & 16.6 & 4.9 & 7.1 & 5.6 & 20.4\\
    PON~\cite{roddick2020predicting}    & 60.4 & 31.0 & 28.0 & 18.4 & 24.7 & 16.8 & 20.8 & 16.6 & 12.3 & 8.2 & 7.0 & 22.2 \\
    LSS~\cite{philion2020lift} & 55.9 & 34.4 & 31.3 & 23.7 & 27.3 & 16.8 & 27.3 & 17.0 & 9.2 & 6.8 & 6.6 & 23.3 \\
    PYVA~\cite{yang2021projecting}   & 56.2 & 32.2 & 26.4 & 21.3 & 19.3 & 13.2 & 21.4 & 12.5 & 7.4& 4.2 & 3.5 & 19.8 \\
    DiffBEV~\cite{zou2024diffbev}  & 65.4 & 41.1 & \textbf{41.3} & 28.4 & 38.9 & 23.1 & 33.7 & 21.1 & 8.4 & 9.6 & 14.4 & 29.6 \\    
    \midrule
    \textbf{Ours} & \textbf{74.7} & \textbf{46.2} & 39.1 & \textbf{39.9} & \textbf{39.8} & \textbf{26.1} & \textbf{37.8} & \textbf{30.9} & \textbf{12.3} & \textbf{16.2} & \textbf{14.7} & \textbf{34.4} \\
    \bottomrule
 \end{tabular}
 }
 \label{tab:multiclass_nuscenes}
\end{table}

%% file: tab/multiclass_argoverse.tex
\begin{table}[htbp] 
 \centering
 \small
 \setlength{\tabcolsep}{2pt} 
 \caption{\textbf{Multi-class segmentation results on Argoverse.}}
 \scalebox{0.8}{
 \begin{tabular}{cccccccccc}
    \toprule
    \textbf{Method} & \textbf{Driv.} & \textbf{Vehicle} & \textbf{Ped.} & \textbf{L. Veh.} & \textbf{Bicycle} & \textbf{Bus} & \textbf{Trailer} & \textbf{Motor.} & \textbf{Mean}\\
    \midrule
    VED~\cite{lu2019monocular}  & 62.9 & 14.0 & 1.0 & 3.9 & 0.0 & 12.3 & 1.3 & 0.0 & 11.9 \\
    VPN~\cite{pan2020cross}  & 64.9 & 23.9 & 6.2 & 9.7 & 0.9 & 3.0 & 0.4 & 1.0 & 13.9 \\
    PON~\cite{roddick2020predicting}  & 65.4 & 31.4 & 7.4 & 11.1 & 3.6 & 11.0 & 0.7 & 5.7 & 17.0 \\
    TaDe~\cite{zhao2024improving} & 68.3 & 34.5 & 9.3 & 14.7 & \textbf{4.4} & 37.8 & 3.1 & 6.4 & 22.3 \\
    \midrule
    \textbf{Ours} & \textbf{77.2} & \textbf{36.4} & \textbf{12.3} & \textbf{15.3} & 3.9 & \textbf{38.6} & \textbf{4.5} & \textbf{6.6} & \textbf{24.4} \\
    \bottomrule
 \end{tabular}
 }
 \label{tab:multiclass_argoverse}
\end{table}

%% file: tab/frozen_backbone_finetune.tex
\begin{table}[h]
    \centering
    \small
    \caption{\textbf{Quantitative assessment (IoU) of our geometry-aligned feature on downstream tasks.}}
    \scalebox{0.9}{
    \begin{tabular}{lccc}
        \toprule
        \textbf{Method} & 
        \textbf{Vehicle (\(\uparrow\))} & 
        \textbf{Pedestrian (\(\uparrow\))} & 
        \textbf{Lane (\(\uparrow\))} \\
        \midrule
        LSS \cite{philion2020lift} & 32.1 & 15.0 & 20.0 \\
        BEVFormer \cite{bevformer} & 35.8 & 16.4 & 25.7 \\
        w/o joint (Ours) & 34.5 & 15.9 & 49.2 \\
        \midrule
        Ours & \textbf{39.6} & \textbf{20.1} & \textbf{60.2} \\
        \bottomrule
    \end{tabular}}
    \label{tab:frozen_backbone_finetuning}
\end{table}

%% file: tab/combined_ablation.tex
\begin{table}[t]
\centering
\small

\begin{minipage}[t]{0.48\linewidth}
\centering
\caption{\textbf{Ablation study on different modules and designs.}}
\label{tab:ablation_module}
\vspace{-0.5em}
{\setlength{\tabcolsep}{5pt}

\begin{tabular}{lccc}
\toprule
\textbf{Method} & \textbf{Vehicle} & \textbf{Ped.} & \textbf{Lane} \\
\midrule
Ours (full) & \textbf{39.6} & \textbf{20.1} & \textbf{60.2} \\
w/o 3-stage & 38.3 & 18.1 & 56.7 \\
w/o DINO & 37.9 & 17.6 & 56.3 \\
\bottomrule
\end{tabular}}
\end{minipage}
\hfill
\begin{minipage}[t]{0.48\linewidth}
\centering
\caption{\textbf{Ablation study on the placement of BEV camera.}}
\label{tab:ablation_height}
\vspace{-0.5em}
{\setlength{\tabcolsep}{5pt}
\begin{tabular}{lccc}
\toprule
\textbf{Height} & 
\textbf{Vehicle} & 
\textbf{Ped. } & 
\textbf{Lane } \\
\midrule
0 m & 37.4 & 17.5 & 58.9 \\
3 m & \textbf{39.6} & \textbf{20.1} & \textbf{60.2} \\
5 m & 39.2 & 19.1 & 59.9 \\
\bottomrule
\end{tabular}}
\label{tab:height_ablation}
\end{minipage}

\end{table}

%% file: sec/5_conclusion.tex
\section{Conclusion}
\label{sec:conclusion}
In this work, we present Splat2BEV, a novel Gaussian Splatting–assisted framework for BEV perception. Unlike existing implicit end-to-end BEV prediction methods,
Splat2BEV incorporates explicit 3D scene reconstruction into the BEV perception pipeline, enabling the learning of geometry-aligned BEV representations.
Comprehensive experiments across multiple downstream tasks demonstrate the effectiveness of our approach and validate the high quality of the learned geometry-aligned features. 